\documentclass[10pt,letterpaper]{article}
\usepackage[top=0.85in,left=2.75in,footskip=0.75in]{geometry}
\usepackage{caption}
\usepackage{subcaption}
\usepackage{graphicx}
\usepackage{amsmath,amssymb}

\usepackage{changepage}

\usepackage{textcomp,marvosym}

\usepackage{cite}

\usepackage{nameref,hyperref}

\usepackage[right]{lineno}

\usepackage[nopatch=eqnum]{microtype}
\DisableLigatures[f]{encoding = *, family = * }

\usepackage[table]{xcolor}

\usepackage{array}

\newcolumntype{+}{!{\vrule width 2pt}}

\newlength\savedwidth

\newcommand\thickhline{\noalign{\global\savedwidth\arrayrulewidth\global\arrayrulewidth 2pt}%
\hline
\noalign{\global\arrayrulewidth\savedwidth}}

\raggedright
\usepackage[aboveskip=1pt,labelfont=bf,labelsep=period,justification=raggedright,singlelinecheck=off]{caption}

\makeatletter
\renewcommand{\@biblabel}[1]{\quad#1.}
\makeatother

\usepackage{lastpage,fancyhdr,graphicx}
\usepackage{epstopdf}
\fancyheadoffset[L]{2.25in}
\fancyfootoffset[L]{2.25in}
\begin{document}
\vspace*{0.2in}

\begin{flushleft}
{\Large
\textbf\newline{A computational framework for longitudinal medication adherence prediction in breast cancer survivors: A social cognitive theory based approach} 
}
\newline
\\
Navreet Kaur\textsuperscript{1*},
Manuel Gonzales, IV\textsuperscript{2},
Cristian Garcia Alcaraz\textsuperscript{2},
Jiaqi Gong\textsuperscript{3},
Kristen J. Wells\textsuperscript{2,4},
Laura E. Barnes\textsuperscript{1}
\\
\bigskip
\textbf{1} Department of Systems and Information Engineering, University of Virginia, Charlottesville, Virginia,  United States of America
\\
\textbf{2} SDSU/UC San Diego Joint Doctoral Program in Clinical Psychology, San Diego, California, United States of America
\\
\textbf{3} Department of Computer Science, The University of Alabama, Tuscaloosa, Alabama, United States of America
\\
\textbf{4} Department of Psychology, San Diego State University, San Diego, California, United States of America
\\
\bigskip

%
%






* nk3xq@virginia.edu

\end{flushleft}
\section*{Abstract}
Non-adherence to medications is a critical concern since nearly half of patients with chronic illnesses do not follow their prescribed medication regimens, leading to increased mortality, costs, and preventable human distress. Amongst stage 0-3 breast cancer survivors, adherence to long-term adjuvant endocrine therapy (i.e., Tamoxifen and aromatase inhibitors) is associated with a significant increase in recurrence-free survival. This work aims to develop multi-scale models of medication adherence to understand the significance of different factors influencing adherence across varying time frames. We introduce a computational framework guided by Social Cognitive Theory for multi-scale (daily and weekly) modeling of longitudinal medication adherence. Our models employ both dynamic medication-taking patterns in the recent past (dynamic factors) as well as less frequently changing factors (static factors) for adherence prediction. Additionally, we assess the significance of various factors in influencing adherence behavior across different time scales. Our models outperform traditional machine learning counterparts in both daily and weekly tasks in terms of both accuracy and specificity. Daily models achieved an accuracy of 87.25\% (Precision – 92.04\%, Recall – 93.15\%, Specificity – 77.50\%), and weekly models, an accuracy of 76.04\% (Precision – 75.83\%, Recall – 85.80\%, Specificity – 72.30\%). Notably, dynamic past medication-taking patterns prove most valuable for predicting daily adherence, while a combination of dynamic and static factors is significant for macro-level weekly adherence patterns. While our models exhibit strong predictive performance, they are constrained by potential cohort-specific biases, reliance on self-reported adherence data, and a limited understanding of the context around non-adherence. Future research will focus on external validation across diverse populations and explore the real-world implementation of sensor-rich systems for a more comprehensive assessment of medication adherence. Nonetheless, we assessed a theory-informed, multi-scale approach to predict adherence, and our findings offer valuable insights to guide the designing of personalized, technology-driven adherence interventions and fostering collaboration among patients, healthcare providers, and caregivers to support long-term adherence.

\section*{Author summary}
Ensuring adherence to medications is crucial for optimal health outcomes. Yet, many individuals face challenges in maintaining consistent adherence rates, resulting in adverse consequences. In our study, we propose an innovative approach to integrate various personal, environmental, and behavioral factors to dynamically forecast the risk of medication non-adherence at daily and weekly time-scale. Personal factors include physiological elements such as side effects, adverse events, etc., as well as psychological factors like self-efficacy, and positive beliefs concerning medications. Environmental factors encompass social, healthcare systems, and physical surroundings. Additionally, behavioral factors such as past medication-taking patterns – weekend vs weekday adherence behavior, morning vs evening medication-taking routine, etc. also play a role. Our findings align with existing literature, indicating that past medication-taking patterns strongly predict daily adherence. However, demographic factors and relatively stable measures like self-efficacy and social support show weaker associations with daily adherence. Conversely, in our weekly models, less frequently changing variables such as perceived susceptibility, decision regret, and self-efficacy, alongside dynamic weekly factors, contribute significantly to explaining adherence patterns at both the macro and micro levels. These insights can enable proactive clinical interventions aimed at optimizing adherence levels.


\section*{Introduction}
In the United States in 2023, an estimated 297,790 women and 2,800 men were diagnosed with invasive breast cancer, along with 55,720 cases of ductal carcinoma in situ (DCIS) in women \cite{refbcstats}. However, with advances in medicine, about 85.5\% of women with breast cancer can survive 5 years, and up to 70\% can survive 10 years after diagnosis if they follow their treatment regimen \cite{allemani2013predictions}. Certain breast cancer survivors (BCS) who are diagnosed with hormone-receptor-positive breast cancer are prescribed endocrine (hormone) therapies (e.g., tamoxifen or aromatase inhibitors) to reduce the chance of recurrence of these cancers by nearly half \cite{bcfacts}.

Non-adherence refers to when patients are unable to follow prescribed medication regimens as intended, including missed doses, incomplete courses, or irregular schedules \cite{adhfactswho}. Even though endocrine therapy (ET) plays a pivotal role in enhancing the long-term survival prospects of individuals with hormone receptor-positive breast cancer, adherence, and persistence rates are often sub-optimal, reducing the effectiveness of the medications \cite{haque2012effectiveness}. Considering that endocrine therapies are typically prescribed to be followed for up to a decade, ensuring consistent adherence over the long term can be demanding. According to a 2017 systematic review, adherence to tamoxifen and aromatase inhibitors ranges from 65\% to 79\% and 72\% to 80\%, respectively, but falls over the course of treatment to ~50\% by the fourth or fifth year \cite{moon2017barriers}.

Medication non-adherence is a complex problem and is associated with a multitude of factors at multiple socio-ecologic levels. While reviews by Stawarz et al. and Nieuwlaat et al. indicate that existing interventions to improve medication adherence often target factors commonly associated with non-adherence across groups of individuals, these strategies may not suit specific individuals given the timing and other contextual factors \cite{stawarz2014don,nieuwlaat2014interventions}. Additionally, existing interventions typically address a limited number of barriers and facilitators at once. Indeed, prior evidence suggests that one-way interventions like patient information and education alone did not significantly impact adherence unless integrated with other components, such as communication strategies \cite{heiney2019systematic}. Several studies in the literature suggest that published interventions have limited success in promoting adherence to adjuvant endocrine therapy (AET) \cite{finitsis2019interventions,hurtado2016behavioral}.

Machine learning advancements enable the analysis of extensive patient data to identify behavioral patterns, showing promise in healthcare applications, including medication adherence. Aziz et al. used various Machine Learning (ML) approaches, such as Random Forests (RF), Artificial Neural Networks (ANN), and Support Vector Regression (SVR) models, to identify factors associated with hypertensive patients' medication adherence levels for the prior month \cite{aziz2020determining}. They identified educational level, marital status, monthly income, and beliefs about medication as significant factors associated with prior month medication adherence.  Zhang et al. used Logistic Regression (LR) models with electronic health records (EHR) and publicly available administrative data from the U.S. Social Security Administration to predict self-reported adherence status over the last three months\cite{zhang2016identifying}. The results showed that the time difference between receiving a social security check and filling a prescription and medication adherence were statistically significantly related among a low-income sample. In the retrospective analysis of e-prescriptions by Kardas et al., a statistically significant association was found between age and medication non-adherence -defined as not filling a prescription within one month of issuing. The older groups showed lower rates of primary non-adherence \cite{kardas2020primary}. Similarly, the authors of \cite{desai2019risk} used LR models with pharmacy claims data over a 1 year period to predict adherence (Medication possession ratios $\geq 80\%$  considered as adherent). The results showed significant associations between 7 predictors (gender, age, race, comorbidity score, medication type, health maintenance organization coverage, and emergency room visit) and adherence.

The prevailing research in the field has primarily relied on a cumulative measure of adherence rather than longitudinal adherence patterns on different time scales (e.g., daily, weekly, monthly, etc.). Commonly used adherence measures in the literature, such as self-reported adherence over extended periods (e.g., adherence levels over the past few months) and pharmacy refill or claims data, provide only broad estimates of medication use. These methods often fail to capture the dynamic nature of adherence behaviors and are susceptible to recall bias, limiting their accuracy and reliability. Moreover, the commonly used predictors of adherence capture attributes or behaviors that are either fixed or vary slowly over time. These factors include, but are not restricted to, socio-demographic variables such as age, gender, and income; clinical attributes such as comorbidities, medication regimens, and laboratory evaluations; and surveys gauging aspects like beliefs about medication\cite{aziz2020determining,haas2019medication,karanasiou2016predicting,kim2019predictors}. We will collectively call them \textbf{static factors} in the rest of this paper. However, all the aforementioned data is recorded at widely spaced intervals, failing to capture the contextual nuances surrounding specific adherence behaviors. Given that human behavior is dynamic and contexts evolve over time, it is imperative to conduct longitudinal monitoring and assessment of medication adherence at different time scales. A few works have highlighted the significance of prior adherence for predicting future adherence. In the works of Gu et al., feature importance demonstrated that historical adherence data contains the most predictive information with respect to adherence \cite{gu2021predicting}. In work by Koesmahargyo et al., results further demonstrate that as predictions become more dynamic (prediction of next week, next day), real-time measurement of dosing increases in importance, and static features (e.g., condition) decrease in importance \cite{koesmahargyo2020accuracy}. We will refer to this as the \textbf{dynamic factors} category, representing factors that fluctuate frequently over time, throughout the remainder of the paper.

Social Cognitive Theory (SCT) \cite{bandura1977social, kelder2015individuals} is a commonly used health behavior theory that evaluates how environmental, personal, and behavioral factors constantly interact with one another through reciprocal determinism. Personal factors include physiological (e.g., side and adverse effects and functional impairment) and psychological factors (e.g., behavioral capacity, self-efficacy, and positive beliefs about medications). Environmental factors include the social, health-system, and physical environment. Behavioral factors, such as eating breakfast at a regular time, can serve as a cue to initiate the behavior of interest \cite{rolling2016effect}. In the field of medication adherence, gaining insights into the interplay of the environmental, personal, and behavioral factors influencing medication-taking will enable the design of personalized interventions to prevent non-adherence. A comprehensive understanding of the context surrounding medication-taking can reveal crucial details about the timing, location, social influences, and surrounding conditions in which adherence or non-adherence occurs. Thus, our research blends SCT to guide the selection of Multi-scale Modeling and Intervention (MMI) constructs with advanced computational models to gain a better understanding of adherence patterns. 

Unlike existing approaches that often examine these factors in isolation or focus on cumulative adherence over extended periods (e.g., several months), our computational framework models the simultaneous interactions among behavioral, environmental, and personal factors and analyzes dynamic longitudinal adherence patterns. Our approach targets multi-scale adherence prediction, including both daily and weekly adherence models that take into account the nature of these variables, the rate at which they evolve/fluctuate as well as the granularity at which they are measured. Our modeling architecture integrates both static and dynamic factors by employing a Long Short-Term Memory (LSTM) layer to capture dynamic, sequential adherence behavior patterns and a Feedforward Neural Network (FNN) to process static/slowly varying personal and environmental characteristics measured by the surveys.  

In this work, our goal is the dynamic prediction of the risk of medication non-adherence that can enable proactive clinical interventions aimed at optimizing adherence. We successfully applied our proposed framework to data from BCS who were prescribed long-term endocrine therapy. The main contributions of this work include 1) multi-scale modeling of dynamic adherence patterns; 2) a novel computational approach for combining data at different time scales for medication adherence prediction; and 3) an understanding of the importance of various factors in determining adherence behaviors.


\section*{Materials and methods}
\label{sec:materials:methods}

\subsection*{Ethical approval}
The study procedures were approved by the Sharp HealthCare Institutional Review Board (IRB \# 1308810) and the SDSU Institutional Review Board (IRB \# 1239088).  Written informed consent was obtained from all participants to participate in the study.
\subsection*{Patient Population and Data Collection}

The data used in this paper were obtained from 36 breast cancer survivors who were part of a randomized controlled trial supported by the National Cancer Institute (R21CA161077) to examine the preliminary efficacy of a patient navigation intervention. This intervention was aimed at medically and historically underserved stage 0-3 hormone-receptor-positive BCS. Fifteen of these 36 patients were assigned to the patient navigation intervention, whereas the remaining received the usual medical care plus a cancer survivorship educational booklet (UC). Participants were recruited at the Cancer Center at Sharp Chula Vista Medical Center at a 6-month post-surgical visit indicating the end of primary treatment that includes surgery, chemotherapy, and radiation; or at the last treatment visit (i.e., chemotherapy) for patients who do not receive surgery or whose radiation or chemotherapy treatment lasts longer than the 6-month post-surgery appointment. The Cancer Center at Sharp Chula Vista Medical Center medical records were reviewed at baseline and 8-month follow-up.

The BCS were eligible if their administrative records indicate they are underserved by 1 or more of the following characteristics: (1) being a member of a historically underserved ethnic or racial minority group (i.e., African-American or Black, Asian-American, Hispanic/Latina, Native American); (2) lacking private health insurance; (3) receiving public health insurance, excluding Medicare (e.g., Medicaid). In addition, participants were included if they: (1) have been diagnosed with HR+ breast cancer in stages 1 to 3 and prescribed AET; (2) have completed all surgery, radiation, and chemotherapy (not including AET) within the past 2 months; (3) speak Spanish or English; (4); are $>$ 18 years of age; (5) live in San Diego County, California; and (6) are able to provide informed consent. Participants were excluded if they were previously diagnosed with any other type of cancer, except for non-melanoma skin cancer.

Table \ref{tab:characteristics} shows the distribution of demographics in the included patient population sample.  

\begin{table}[!ht]
\begin{adjustwidth}{-2.25in}{0in} 
\centering
\caption{
{\bf Characteristics of study participants (n =32).}}
\begin{tabular}{|l|l|}
\hline
{\bf Characteristic} & {\bf Number (Percent)} \\ \thickhline

\textbf{Age (years)} & \\
Mean (SD) & 51.63 (8.85)\\ \hline
\textbf{Gender} & \\
Female & 32 (100\%)\\ \hline
\textbf{Race} &  \\
White  & 14 (43.75\%) \\ 
Black/African American & 4 (12.50\%)\\
Asian   & 2 (6.25\%) \\
Native Hawaiian/Other Pacific Islander  & 1 (3.13\%) \\
Native American/Alaskan Native  & 1 (3.13\%) \\
Unknown  & 10 (31.25\% )\\ \hline
\textbf{Latin(o/e/x)} &  \\
Latin(o/a/e/x)  & 25 (78.13\%) \\
Not Latin(o/a/e/x) & 7 (21.88\%)\\ \hline
\textbf{Stage of Cancer} & \\ 
DCIS & 5 (15.63\%)\\ 
Early breast cancer & 25 (78.13\%)\\ 
Locally advanced breast cancer & 2 (6.25\%)\\  \hline
\textbf{Preferred Language} & \\ 
English & 7 (21.88\%)\\ 
Spanish & 24 (75\%)\\
Other & 1 (3.13\%)\\ \hline
\textbf{Intervention Arm} & \\ 
Patient Navigation intervention & 15 (46.88\%)\\ 
Usual Care & 17 (53.13\%)\\ \hline
\end{tabular}
\label{tab:characteristics}
\end{adjustwidth}
\end{table}

Within two months following completion of surgery, chemotherapy, and/or radiation, participants were given a smart pill dispenser called a medication event monitoring system (MEMS) to track their intake of AET. MEMS devices recorded the time of bottle opening, which represented medication-taking events, every day for eight months. The data regarding factors that potentially influence the patients' medication-taking behavior were collected in-person via paper surveys at baseline (0 month), 4-month, and 8-month follow-ups. The instruments used in the survey are listed in the Supporting Information \nameref{S1_Appendix}, along with a description of the constructs they measure and the scoring methods. These instruments broadly measure various personal and environmental factors that are associated with medication-taking behaviors. 

Out of the 36 patients enrolled in the study, we excluded four patients and used the data from the remaining 32. Amongst the excluded patients, two patients had completely missed surveys at the 4-month follow-up, while all of them had missed responding to surveys administered at 8 months. Three of these four participants did not have any data from the smart pill device, which could have been a result of the device being faulty, patients not being comfortable with the device, patients losing the device, or other unknown reasons. Since the smart pill device data serves as the ground truth for adherence status, its absence in three participants rendered their data non-viable, necessitating their exclusion from the analysis. This type of missingness is completely at random and unexplained by observed and unobserved data. Thus, the removal of these participants' data will not alter the results of our models. The fourth excluded participant missed both the 4-month and 8-month follow-up visits and had 100\% missing survey data for these time points (due to early dropout from the study), with an overall survey data availability of only 33.33\% (since the baseline was answered). Given that survey data is a crucial component of our models, this participant was also excluded.
Section \nameref{ssec : feature_engg} provides a detailed description of the features extracted from both the MEMS device and survey data. To ensure data reliability, we applied a 60\% availability threshold for including a feature in our models.

\subsection*{Multi-scale Adherence}
\begin{figure}[!h]
\includegraphics[width=\linewidth]{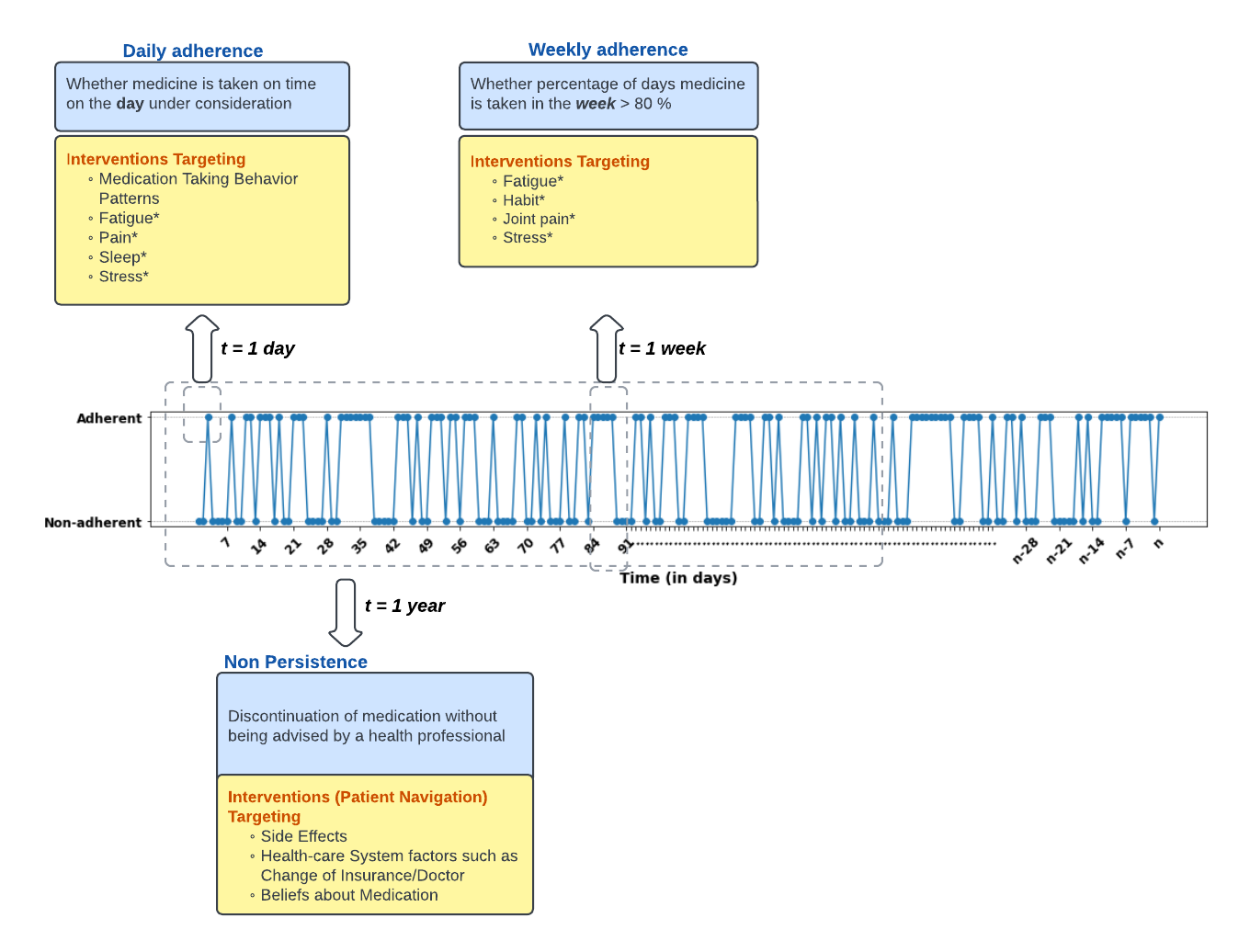}
\caption{ \bf Multi-scale modeling and intervention for medication adherence}
(*indicates the corresponding factor is intervened upon at more than one time scale)
\label{fig:fig1}
\end{figure}

Medication adherence can be examined at different time scales, such as daily, weekly, and yearly, as shown in Fig \ref{fig:fig1}, to provide insights into how patients' medication-taking behaviors evolve over time. The use of electronic adherence monitors has made longitudinal adherence monitoring possible. By tracking adherence on a daily basis, we gain a granular understanding of short-term fluctuations, while weekly patterns reveal relatively broader trends. Another type of non-adherence called non-persistence might occur over an extended period of time (for e.g., a year) in which patients decide to stop taking a medication after starting it, without being advised by a health professional to do so. It is important to note that factors associated with adherence at different time scales might not always be mutually exclusive. Daily adherence may be associated with factors such as medication-taking behavior patterns, fatigue, pain, sleep, stress, etc., which can be the focus of interventions targeting daily adherence. Similarly, weekly adherence may be associated with factors such as fatigue, habit, joint pain, stress, etc., which can be the focus of interventions targeting weekly adherence. Non-persistence can be potentially associated with factors such as side effects, change of insurance or doctor, beliefs in medicine etc. Moreover, the idea is that patients will apply the knowledge gained from daily and weekly adherence interventions, ultimately contributing to an enhancement in persistence. This comprehensive perspective can enable healthcare providers to tailor interventions that address both immediate and long-term adherence challenges. In addition, there might be other factors that are time-independent, time-invariant and typically non-modifiable, such as demographics and personality traits, which are not useful in terms of intervention targets. However, they can provide insights into subgroups at higher risk for medication non-adherence. 

In this work, BCS were recommended to take one AET pill once per day, and we considered the problem of prediction of daily adherence and daily adherence over a week's time (referred to as weekly adherence in the rest of the paper). Daily medication adherence was defined as taking the medication within 24±6  hours of taking the last dose\cite{zeller2008patients}. Thus, if the time difference between two consecutive bottle-opening events is within 24±6 hours, daily adherence is considered positive; otherwise, it is negative. Weekly adherence was defined as following a treatment regimen more than 80\% of the time. Even though there is no gold standard threshold for optimal adherence to AET, many works in literature have used a threshold of 80\%  \cite{sackett1975randomised,bhatta2013factors}.

\subsection*{Feature Engineering}
\label{ssec : feature_engg}
\begin{figure}[!h]
\includegraphics[width=\linewidth]{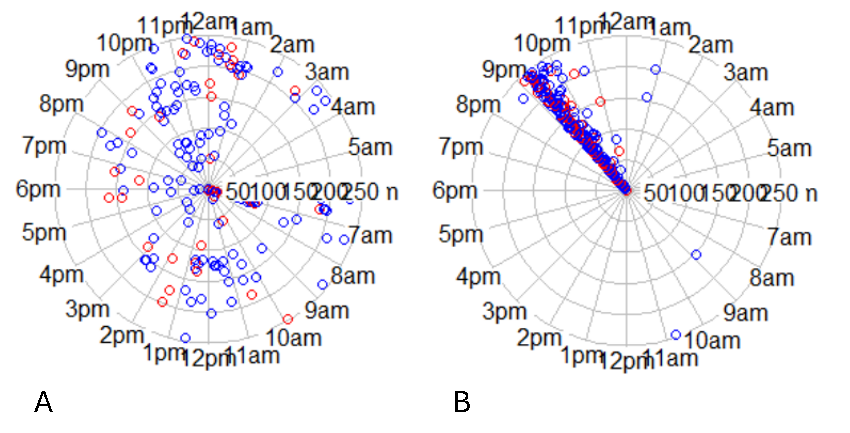}
\caption{ \bf Dynamic adherence patterns: A) patient 53, B) patient 65.}
Patient 53 has irregular adherence behavior with a large variation in medication-taking time over the course of the study; Patient 65 follows a regular schedule by taking medication between 9 pm - 10 pm during the majority of the days.
\label{fig:fig2}
\end{figure}

\begin{table}[!ht]
\begin{adjustwidth}{-2.25in}{0in}
\centering
\caption{\bf Personal, environmental, and behavioral constructs and their measurement.} \label{tab:sct_cons}
 \begin{tabular}{p{2.6cm}p{5cm}p{9cm}} 
 \hline
\textbf{SCT Construct} & \textbf{Data} & \textbf{Description} \\ \thickhline
 \hline\hline
 & Communication and Attitudinal Self-Efficacy (CASE) Scale \cite{aubel2019patient} & A 12-item measure of self-efficacy in coping with cancer and related health services\\
 & Pearlin Mastery Scale \cite{lim2022psychometrics} & A 7-item measure of sense of mastery and control \\
 & Perceived Stress Scale \cite{cohen1983global} & A 14-item measure of perceived stress\\
 & FACT-G Scale \cite{yost2013functional} & A 27-item measure of quality of life comprising physical well-being, emotional well-being, functional well-being, and social well-being (also included in environmental factors)\\
 & FACIT-SP Scale \cite{facitsp} & A 12-item measure of spirituality with chronic and/or life-threatening illness\\
 \textbf{Personal}& MD Anderson Symptom Inventory \cite{sailors2013validating} & A 19-item measure of symptom (e.g., pain, fatigue, etc.) severity and interference with life\\
 & Breast Cancer Symptom Prevention Trial Checklist \cite{alfano2006psychometric} & A 43-item measure of symptoms associated with endocrine therapy.\\
 & Decision Regret Scale \cite{brehaut2003validation} & A 5-item measure of  distress or remorse regarding the decision to take endocrine therapy \\
 & Perceived Susceptibility Scale \cite{champion1999revised} & A 3-item measure of perceived likelihood of breast cancer recurrence\\
 & Medication Adherence Self Efficacy Scale \cite{ogedegbe2003development} & A 42-item measure of self-efficacy in adherence to prescribed anti-hormonal medications\\
 & Beliefs about Medication Questionnaire \cite{horne1999beliefs} & A 18-item measure of general beliefs (Overuse and Harm) and specific beliefs (Necessity and Concerns) about medication\\

 \hline
 & Kraus and Borawski-Clark Social Support Scale \cite{perkins2007individual} & A 14-item measure of the quantity of social support and satisfaction with it\\
 & FACT-G Scale & A 27-item measure of the quality of life comprising physical well-being, emotional well-being, functional well-being, and social well-being (also included in personal factors)\\
 \textbf{Environmental}& Barriers to Care Scale  \cite{heckman1998barriers} & A 28-item measure of barriers related to geography/distance, medical and psychological issues, community stigma, and healthcare and personal resources\\
 & Patient Satisfaction with Cancer Care Scale \cite{jean2011structural} & An 17-item measure of patient satisfaction with cancer care \\
 \hline
 & Daily routines and recent adherence patterns from MEMS device & Adherence over recent days\\
& &Time of medication taking (Morning, Afternoon, Evening, Night)\\
& & Weekday vs weekend medication patterns\\
\textbf{Behavioral}& Weekly routines and recent adherence patterns from MEMS device & Adherence over recent weeks\\
& & Weekend adherence levels (adherent on 0 or 1 or 2 weekend days)\\
& & Most frequent medication-taking time-epoch in the week (Morning, Afternoon, Evening, Night)\\
& & Mean hour of medication taking in the week\\
& & Standard deviation of hour of medication-taking in the week.\\[1ex]
\hline
 \end{tabular}
 \end{adjustwidth}
\end{table}
SCT served as the foundation for the selection of multi-scale modeling features measuring personal, behavioral, and environmental factors potentially influencing adherence behaviors. By leveraging SCT’s framework, we identified key determinants that shape adherence patterns, ensuring a comprehensive and theoretically grounded approach. Table \ref{tab:sct_cons} outlines the specific factors included in our analysis and illustrates how they align with the core components of SCT.
Our analysis consisted of predicting medication adherence using two types of features- dynamic features and static features, which are listed in Table \ref{tab:sct_cons}. The dynamic features refer to the factors that keep changing over time, such as daily or weekly medication-taking behavioral patterns. These were extracted from the timestamped MEMS records. The daily dynamic features included: 1) whether the patient was adherent on a day in the recent past ($t-i \ is\_adherent$); 2) timing of bottle opening in the recent past and ($t- i \ Morning$, $t- i \ Afternoon$, $t- i \ Evening$, $t- i \ Night$); 3) whether a particular dosing day in the recent past was weekend or weekday ($t-i \ is\_Weekend$).  The definition of the recent past in our analysis is 7 days prior to the target day, and thus $i \in \{1,2,3,4,5,6,7\}$. We tested various lag values ranging from 4 to 14 days and found that the model with 7-day lagged features produced the best results. Therefore, we present the findings of the 7-day lagged daily adherence model in this paper. The choice of the included features were informed both by literature as well of exploration of the data at hand. We know from the literature that past behavior is highly predictive of future behavior \cite{gu2021predicting,koesmahargyo2020accuracy}. As we can see from Fig \ref{fig:fig2}, study participants varied widely in their medication-taking habits. Participant 65 follows a consistent routine and takes medicine at nearly the same time every day. On the contrary, participant 53 follows a rather irregular pattern over time with large variations in the timing of medicine intake. The timing of bottle opening was converted into four time epochs - Morning (6 am-11:59 am), Afternoon (12 pm-5:59 pm), Evening (6 pm-11:59 pm) and Night (12 am-5:59 am). As we can see the difference in weekend (shown in red) vs weekday (shown in blue) patterns of medication taking in Fig \ref{fig:fig2}, the day of the week was converted to weekend vs weekday feature. 

The weekly dynamic features included: 1) weekend adherence levels (adherent on 0/1/2 weekend days; $t-j \ weekend\_adh\_0$ means non-adherent on both Saturday and Sunday, $t-j \ weekend\_adh\_50$ means adherent on either Saturday or Sunday, $t-j \ weekend\_adh\_100$ means adherent on both Saturday and Sunday); 2) most frequent medication-taking time-epoch in the week ($t- j \ Morning$, $t- j \ Afternoon$, $t- j \ Evening$, $t- j \ Night$); 3) mean hour of medication taking in the week ($t-j \  time\_mean$); and 4) standard deviation of hour of medication taking in the week ($t-j \  time\_std$), where $j \in \{1,2,3,4\}$. We tested various lag values ranging from 2 to 6 weeks and found that the model with 4-week lagged features produced the best results. Therefore, we present the findings of the 4-week lagged weekly adherence model in this paper.


The static data refer to factors that stay relatively consistent over a long period of time (e.g. over months) and that corresponds to survey data in our study. The cumulative scores of scales or sub-scales (if applicable) are calculated according to the scoring schemes used in literature and can be found in Supporting Information \nameref{S1_Appendix}, along with the detailed description of each survey. The surveys included were selected based in the domain expertise of the clinical psychology members on the team. Internal consistency was also tested for these surveys before inclusion in our analysis using a cut-off of 0.7 for Cronbach's alpha. Table \ref{tab:sct_cons} also provides information on the type of construct each feature represents (e.g., personal, environmental, behavioral).

\subsection*{Data Processing}
\label{data_processing}
As described above, only 32 of the 36 patients were included in our analysis due to missingness of data. We analyzed patients' medication-taking data for unusual patterns at the start of data collection. During this period, many participants exhibited either exceptionally high adherence or signs of adjustment to the MEMS device, such as missing data for an entire week following the start of data collection. Thus, the first month of the MEMS device data was excluded from the analysis to avoid the impact of the Hawthorne effect on our findings \cite{sedgwick2015understanding}. The remaining data still had missing values both in survey responses and smart pill dispenser records.

To ensure data reliability, we applied a 60\% availability threshold for including a feature in our models and performed imputation for the missing data. For the numeric survey data, we imputed the missing values using sklearn's Iterative Imputer, which uses a multivariate imputation technique that estimates each feature from all the others \cite{pedregosa2011scikit}. Iterative imputation considers and preserves the relationship between different features as well as provides more accurate imputations when features are correlated. For categorical data, the mode values were imputed and then converted to one hot encoding form which is compatible with the models. In order to prevent the difference in the range of values of different features from impacting the results, we normalized the numeric data using sklearn's MinMaxscaler, which transforms the data by scaling each feature to a given range \cite{pedregosa2011scikit}. Also, since there was a major imbalance in the adherent/non-adherent classes, we used Synthetic Minority Over-sampling Technique (SMOTE) from Python's imblearn package \cite{chawla2002smote}. Following this, the training data underwent feature selection, and subsequently, only the chosen features were extracted and used for both the training and test sets. This is explained in detail in the Section \nameref{sssec:analysis_pipeline}.

\subsection*{Computational Models}

\begin{figure}[!h]
\includegraphics[width=\linewidth]{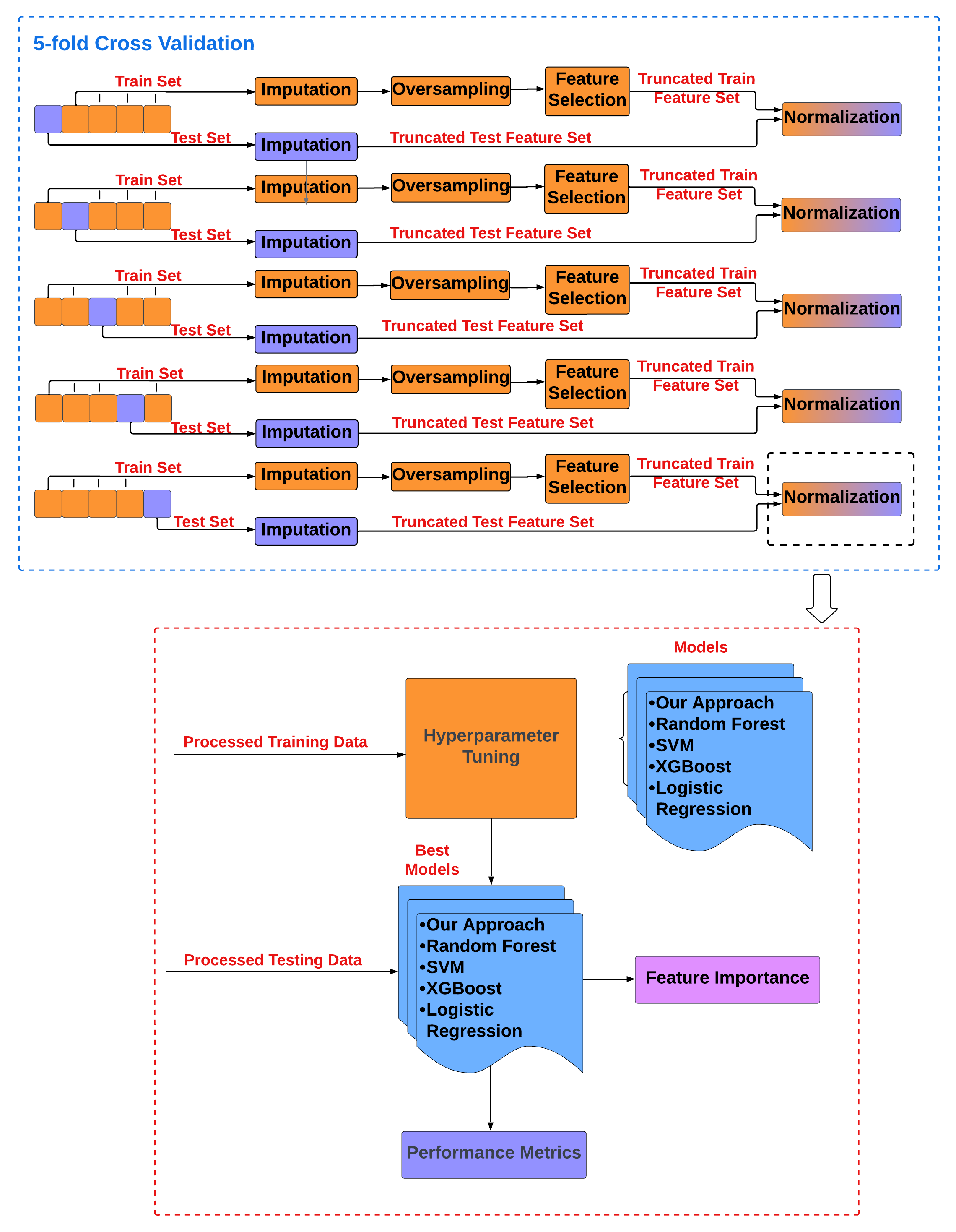}
\caption{ \bf Machine learning pipeline.}
5-fold cross-validation with hyperparameter tuning for model development and evaluation.
\label{fig:fig3}
\end{figure}
\subsubsection*{Analysis Pipeline}\label{sssec:analysis_pipeline}
The step-by-step implementation of our methodology is shown in Fig \ref{fig:fig3}, which is applicable in both the daily and weekly tasks. In addition to our proposed modeling approach described in Section \nameref{model_architechture}, various machine learning models, including random forest, Support Vector Machine (SVM), XGboost, and logistic regression, were built and tested. In order to provide generalizable results, we followed a nested 5-fold cross-validation approach with hyperparameter tuning. The data are split at the subject level, which guards against data leakage by ensuring that subjects who appear in the training set do not appear in the test set. Furthermore, all the data processing steps described above in Section \nameref{data_processing} were carried out separately for train and test sets to avoid any data leakage.

The machine learning models were developed using scikit-learn and xgboost Python packages \cite{pedregosa2011scikit,brownlee2016xgboost} except for the LSTM model, which was built using Tensorflow \cite{abadi2016tensorflow}. The input to the LSTM model has the form \textit{Number of samples * Number of time steps * Number of features}, implying all features that are input into the model have the same number of time steps for each sample. Thus, we included all time steps of any dynamic feature that was selected at least at one lag, resulting in an expanded feature set. For instance, if the feature '$is\_weekend$' was selected only at time $t-1$ in the feature selection process, we included it for all times lags $t-1$ through $t-7$ for that feature. To ensure a fair comparison between the machine learning models random forest, SVM, XGboost, logistic regression, and our approach, the same expanded feature set was used for each model. We performed hyperparameter tuning for all built models. The standard machine learning models were tuned with Ray Tune's TuneGridSearchCV, which performs an exhaustive search over specified parameter values for an estimator \cite{liaw2018tune}. The parameters of the estimator used to apply these methods are optimized by cross-validated grid-search over a parameter grid. For our proposed approach, we used the Keras GridSearch tuner to optimize our models over the whole hyperparameter space \cite{JMLR:v18:16-558}. The models were fitted with the best performing hyperparameters in each fold, and the mean and standard deviation of the performance metrics across different folds are reported.

\begin{figure}[!h]
\includegraphics[width=\linewidth]{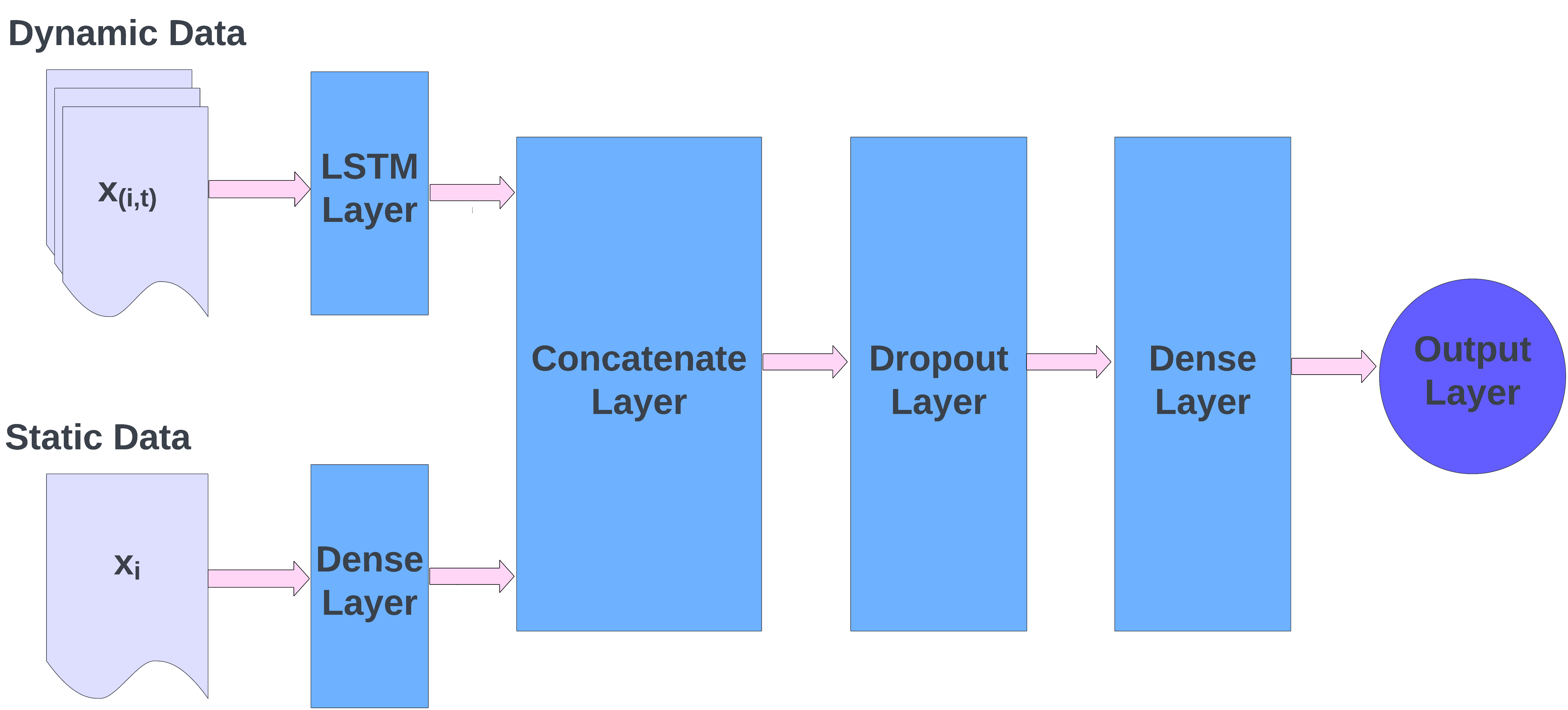}
\caption{ \bf Proposed model architecture to integrate dynamic and static data.}
The model processes static and dynamic data separately. The dynamic part uses an LSTM layer, while the static data is fed into a Feedforward Neural Network (FNN) layer. The hidden states from both the LSTM and FNN layers are concatenated, followed by a dropout layer, a final hidden layer, and an output layer.
\label{fig:fig4}
\end{figure}

\subsubsection*{Modeling Architecture}
\label{model_architechture}

Our modeling approach integrates dynamic and static as shown in the model architecture (Fig \ref{fig:fig4}). The static features, which remain constant over time, provide foundational context, while dynamic features capture temporal variations. Considering the difference in the nature of the static and dynamic features, the underlying constructs they measure, and the time scales at which they are recorded, these features are processed in different ways before being merged into the final predictive model. The model architecture is shown in Fig \ref{fig:fig4}. As we can see, the static and dynamic parts of the data are input separately into the model. The key idea is to use an LSTM layer, which is designed to capture long-term dependencies in sequential data for the dynamic data. On the other hand, static data are processed through a Feedforward Neural Network (FNN) Layer before fusion with the dynamic data. Further, we adopt the concatenation operation to combine the hidden states of both the LSTM and the  Feedforward Neural Network layers. Thus, in our approach, both dynamic and static features are incorporated into the model through concatenation, and the concatenated hidden state encoding both static and dynamic information is followed by a dropout layer, final hidden layer, and output layer with daily/weekly adherence status prediction.

We do not explicitly assign predefined weights to different features, their importance is learned automatically by the model during training. To analyze feature importance, we conducted post-hoc interpretability assessments using SHAP (Shapley Additive Explanations), which is discussed in the \nameref{ssec:featimportance} section. Since the model already achieved strong performance, we did not experiment with explicit feature weighting. However, future research could explore different weighting techniques, such as adjusting loss weights, applying feature scaling strategies, or incorporating domain-driven weight adjustments. In our current work, the model effectively adapted its internal parameters to prioritize the most relevant features without manual intervention, resulting in robust performance.

To mitigate overfitting in our models, we incorporated several techniques, including a dropout layer, early stopping, and input normalization. We used a dropout layer for regularization as it deactivates several neurons randomly at each training step to avoid overfitting. We also implemented an early stopping criterion, which halted training when the optimization objective showed no improvement for five consecutive epochs. This prevented unnecessary training beyond the optimal point, reducing the risk of overfitting. Additionally, as discussed in Section \nameref{data_processing}, we normalized the input features to accelerate and stabilize the learning process. Without normalization, features with large value ranges could lead to disproportionate weight updates during backpropagation, resulting in oscillations, increased variance, and ultimately overfitting.


\section*{Results}
We investigated the performance of several machine learning models, including random forest, SVM, XGBoost, and logistic regression, and compared them with our approach for the daily and weekly prediction tasks. Considering the distinct nature of predictors, our approach introduced a novel method for integrating data across varied time scales as opposed to traditional ML models. Tables \ref{tab:performance} and \ref{tab:weekperformance} show the mean and standard deviation of various performance metrics across all folds of different models. Table \ref{tab:performance} shows our approach demonstrated superior performance, achieving 87.25\% accuracy in daily predictions, particularly excelling in specificity (it is more important to correctly identify cases of non-adherence, thus, specificity is an important metric) with 77.50\%. For the weekly adherence task, our approach consistently outperformed other models in accuracy, recall, and specificity (Table \ref{tab:weekperformance}). 

\begin{table}[!ht]
\begin{adjustwidth}{-2.25in}{0in} 
\centering
\caption{
{\bf Performance comparisons of daily adherence models.}}
\begin{tabular}{|l|l|l|l|l|}
\hline
{\bf Model} & {\bf Accuracy (\%)} & {\bf Precision (\%)} & {\bf Recall (\%)} & {\bf Specificity (\%)}\\ \thickhline
Our Approach & \textbf{87.25 ± 0.29} & \textbf{92.04 ± 0.17} & \textbf{93.15 ± 0.51} & \textbf{77.50 ± 1.20}\\ \hline
Random Forest &  84.64 ± 0.41 & 90.55 ± 0.20 & 91.73 ± 0.43& 74.83 + 0.60\\ \hline
SVM & 85.10 ± 0.62 & 90.02 ± 0.26 & 91.79 ± 0.62 & 72.28 + 0.42\\ \hline
 XGBoost & 86.70 ± 0.15 & 89.73 ± 0.14& 92.34 ± 0.20& 73.97 ± 0.51\\ \hline
Logistic Regression & 82.36 ± 0.32 & 87.01 ± 0.13& 88.54 ± 0.54& 71.44 ± 0.35\\ \hline
\end{tabular}
\begin{flushleft} Our approach outperforms other machine learning models in terms of all reported performance metrics for daily adherence prediction.
\end{flushleft}
\label{tab:performance}
\end{adjustwidth}
\end{table}


\begin{table}[!ht]
\begin{adjustwidth}{-2.25in}{0in} 
\centering
\caption{
{\bf Performance comparisons of weekly adherence models.}}
\begin{tabular}{|l|l|l|l|l|}
\hline
{\bf Model} & {\bf Accuracy (\%)} & {\bf Precision (\%)} & {\bf Recall (\%)} & {\bf Specificity (\%)}\\ \thickhline
Our Approach & \textbf{76.04 ± 0.85} & 75.83 ± 0.82 & \textbf{85.80± 0.92} & \textbf{72.30 ± 1.28}\\ \hline
Random Forest &  75.14 ± 0.67 & 76.04 ± 0.77 & 81.47 ± 1.35 & 67.58 ± 1.00\\ \hline
SVM & 75.10 ± 0.69 & 76.53 ± 1.36 & 80.68 ± 0.91 & 68.46 ± 1.52\\ \hline
XGBoost & 75.05 ± 1.15& 76.35 ± 1.09 & 79.34 ± 1.47 & 67.79 ± 1.31\\ \hline
Logistic Regression & 74.60 ± 0.39 & \textbf{77.21 ± 1.01} & 77.64 ± 1.01 & 71.21 ± 0.92\\ \hline
\end{tabular}
\begin{flushleft} Our approach outperforms other machine learning models in terms of accuracy, recall, and specificity for weekly adherence prediction.
\end{flushleft}
\label{tab:weekperformance}
\end{adjustwidth}
\end{table}

\subsection*{Feature Importance}
\label{ssec:featimportance}
\begin{figure}[!h]
\includegraphics[width=\linewidth]{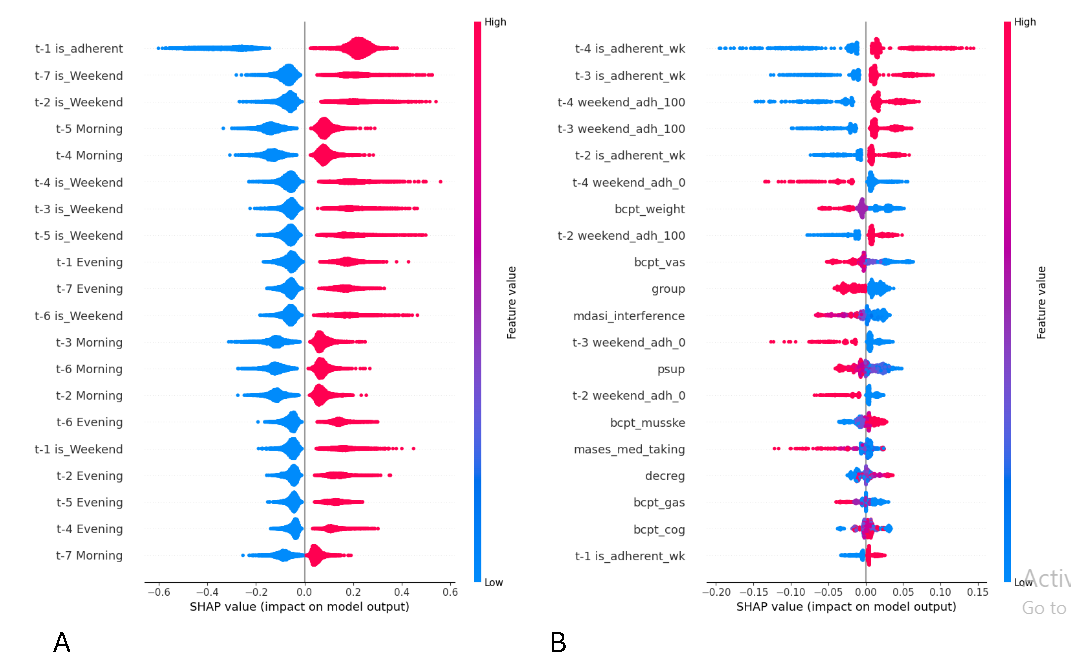}

\caption{{\bf Model interpretation with SHAP for: A) daily medication adherence, B) weekly medication adherence.} \label{fig:fig5}
The y-axis reflects the relative importance of the top 20 features for each task by the vertical order of features (top being the most important and bottom being the least). The x-axis reflects the mean SHAP values of each feature and represents the magnitude and direction of impact of each feature on the output (whether there's a higher/lower chance of taking (right of 0) or not taking (left of 0) medicine). The color bar represents lower/higher values of features and their relative association with adherence/non-adherence.}
\end{figure}

We used the Shapley additive explanation (SHAP) values from the gradient explainer to understand the most important factors associated with medication adherence. The Gradient Explainer combines the ideas from Integrated Gradients, SHAP, and Smooth Grad into a single expected equation. As anticipated, the dynamic features were important in the daily prediction task, whereas a combination of dynamic and static features were significant in the weekly task. A few observations from Fig \ref{fig:fig5} are as follows.

If a person was adherent yesterday ($t-1 \ is\_adherent$), she was more likely to be adherent today and vice versa. The next key factor is $t-7 \ is\_Weekend$, implying that a patient was more likely to be non-adherent if the predicted day fell on a weekend ($t-7 \ is\_Weekend$ denotes it was a weekend seven days ago, and hence, today will be Saturday/Sunday) and vice versa. Similarly, if it was a weekend two days ago ($t-2 \ is\_Weekend$), implying today being Monday/Tuesday, a person was more likely to be non-adherent today. We expected a difference in medication-taking behavior over the weekends (Saturday/Sunday) vs the weekdays (Monday through Friday). However, we found the $is\_Weekend$ feature to be significant at various time lags, including four, three, five, six, and one day ago, in decreasing order of importance. We anticipate this happening due to variations in individual schedules- where different patients may observe free days during different days of the week. We also found that following a morning/evening medication-taking routine in the last week ($t-i \ Morning$ or $t-i \ Evening$, where $i \in \{1,2,3,4,5,6,7\}$) to be associated with a higher likelihood of adherence on the current day. More specifically, if a person took her medicine during the morning or evening time during the last seven days leading up to the predicted day, she is more likely to do so today as well.

In the weekly models, being adherent to medications in the previous weeks was found to be strongly associated with a higher likelihood of adherence in the current week. The adherence in the week four weeks ago \textit{($t-4 \ is\_adherent\_wk$)} was the most significant predictor of current weekly adherence, followed by adherence in the week three weeks ago ($t-3 \ is\_adherent\_wk$) and two weeks ago ($t-2 \ is\_adherent\_wk$). Higher adherence over the weekends in the past weeks was also associated with a higher likelihood of adherence in the current week and vice versa. Specifically, the features $t-4 \ weekend\_adh\_100$, $t-3 \ weekend\_adh\_100$, and $t-2 \ weekend\_adh\_100$, which denoted 100\% adherence over weekends in the weeks that were four, three, and two weeks ago, respectively- were associated with higher adherence in the current week. On the contrary, complete non-adherence on both Saturday and Sunday in the week four weeks ago ($t-4 \ weekend\_adh\_0$) was associated with higher chances of non-adherence in the current week. Besides the past adherence patterns, symptom burden also played a role in determining weekly adherence status. If someone experienced weight-related symptoms ($bcpt\_weight$), vasomotor symptoms ($bcpt\_vas$), or symptom interference with normal activities ($mdasi\_interference$) from the AET, she was more likely to exhibit non-adherence. Patient navigation arm patients ($group$) were more likely to be adherent.  Some additional factors associated with weekly medication adherence, though less influential than the abovementioned ones, included $t-3 \ weekend\_adh\_0$ and $t-2 \ weekend\_adh\_0$, denoting 0\% adherence over weekends in the weeks three and two weeks ago, respectively. They showed an association with lower adherence in the current week. 

Other variables related to weekly medication adherence were perceived susceptibility ($psup$), decision regret ($decreg$),  medication-taking self-efficacy ($mases\_med\_taking$), musculoskeletal pain ($bcpt\_musske$), gastrointestinal symptoms ($bcpt\_gas$), cognitive symptoms ($bcpt\_cog$), and adherence status in the prior week ($t-1 \ is\_adherent\_wk$). Perceived susceptibility to cancer recurrence was associated with higher adherence. Decision regret after choosing to take AET was higher in cases of adherence compared to non-adherence. One possible explanation is that individuals who take their medication may experience more side effects or doubts, leading to feelings of remorse. Amongst the symptom-related features, the occurrence of gastrointestinal symptoms was associated with non-adherence. However, surprisingly, the opposite trend was observed for musculoskeletal and cognitive symptoms, though these relationships were not highly significant. Adherence in the previous week was strongly correlated with adherence in the current week, reinforcing the pattern of consistent medication-taking behavior. Additionally, lower medication-taking self-efficacy (higher values on the scale indicate low self-efficacy) was associated with a greater likelihood of weekly adherence.

While SHAP values provide valuable insights into feature importance, their application in deep learning models comes with certain limitations. One key challenge is that SHAP explanations can be sensitive to correlated features, making it difficult to disentangle individual contributions. Additionally, in highly complex models, SHAP values may not always provide fully intuitive explanations, as they rely on approximations of the model's decision process. Computational cost is another factor, as calculating SHAP values for deep networks can be resource-intensive. Despite these limitations, SHAP remains a useful tool for understanding feature influence.

\section*{Discussion}
In this work, we evaluated theory-informed models for predicting daily and weekly adherence. The existing methods primarily focus on cumulative measures of adherence \cite{aziz2020determining, desai2019risk} as opposed to longitudinal prediction done in a few works \cite{gu2021predicting,koesmahargyo2020accuracy}. In contrast, we adopt a multi-scale approach to adherence prediction, which is essential for informing effective, more personalized interventions. This can enable the design of technology-mediated solutions targeting non-adherence. Additionally, intervention approaches may engage stakeholders, including healthcare providers and family members, in assisting patients at high risk of non-adherence or non-persistence.

We learned, not surprisingly and corroborated by literature \cite{gu2021predicting,koesmahargyo2020accuracy}, that past medication-taking patterns are highly predictive of daily medication-taking. Demographic factors and infrequently changing measures, such as self-efficacy and social support, were not strongly associated with daily medication adherence, corroborating findings in literature \cite{gu2021predicting,koesmahargyo2020accuracy}. However, in our weekly models, less frequently changing factors such as perceived susceptibility \cite{champion1999revised}, decision regret \cite{brehaut2003validation}, and self-efficacy \cite{ogedegbe2003development}, alongside dynamic weekly factors, contribute to explaining macro-level and micro-level adherence patterns, respectively.

These multi-scale models are two lenses into the unique medication-taking experience and differing reasons for non-adherence and can be leveraged in intervention design. For example, in targeting daily adherence, we can leverage past temporal data to tailor an intervention for individuals deviating from their usual morning/evening schedule by sending them a text message.  At the macro or weekly level scale, we can develop interventions that are focused on assisting patients to improve their self-efficacy. In other words, the intervention approaches could be personalized and attempting to improve medication through multiple mechanisms.

While these models are useful, we hypothesize that dynamically changing constructs like stress, sleep, fatigue, and pain, which are more difficult to capture, may also be important for understanding medication-taking behavior.  Ubiquitous devices, such as smartphones and wearables, are a vehicle to provide this more fine-grained, longitudinal data related to medication adherence.  However, existing technology-based interventions focus on cognitive reasons for non-adherence to medications experienced by some people (e.g., forgetting) but fail to account for interactions between cognitive and other types of factors (i.e., environmental, behavioral) that contribute to adherence.  We expect that uncovering the nuances in day-to-day medication adherence patterns, leveraging ubiquitous technologies such as smartphones and wearables to provide the context, could lead to more personalized interventions than currently available. 

To develop a comprehensive approach for capturing the contextual nuances of adherence behaviors, we have designed a Multiscale Modeling and Intervention (MMI) system to support BCS in maintaining adherence to daily oral ET. To develop the MMI system, we conducted usability interviews with 25 BCS on ET, reviewed commercial wrist-worn sensors and medication event monitoring system (MEMS) devices, analyzed ET adherence data from 32 BCS, and conducted a review of the relevant literature. Usability interviews were recorded, transcribed, and analyzed, revealing that participants were open to using an EMA smartphone app, a smartwatch with a companion app, a smart pill bottle or pillbox, and the full MMI system for six months. We reviewed 26 wearable sensors and 18 MEMS devices, selecting the Fitbit Sense and RxCap MEMS for inclusion in the system. Additionally, 32 adherence-related constructs were identified for assessment at baseline, 3 months, and 6 months.

We deployed the MMI system with 20 BCS and are developing computational models to analyze adherence patterns in the collected multi-modal data. This research integrates theory, data-driven modeling, and participant feedback to inform future ET adherence interventions. We are developing smartphone-based intervention modules triggered at key decision points based on computational models. For example, a reminder module will help users with irregular medication habits establish a consistent routine. Another module will engage family, friends, and healthcare providers to offer informational, instrumental, and emotional support for those struggling with adherence. By leveraging multi-modal data to inform personalized interventions, the MMI system aims to enhance ET adherence and improve patient outcomes.

While our findings demonstrate the potential of the MMI system for improving ET adherence, implementing this technology-based solution in resource-limited settings, such as lower- and middle-income countries (LMICs), presents several challenges. Limited access to smartphones, wearables, and stable internet connectivity may hinder widespread adoption. Additionally, variations in digital literacy and healthcare infrastructure could impact the feasibility of integrating such interventions into routine care. The cost of MEMS devices and wearable sensors may also be prohibitive for large-scale deployment in LMICs, necessitating the exploration of more affordable alternatives, such as SMS-based reminders or simplified mobile applications. Future research should investigate culturally and contextually appropriate adaptations of the MMI system, including low-cost, non-smartphone-based interventions and community-driven support models that leverage existing healthcare resources. Collaborating with local stakeholders, including healthcare providers and patient advocacy groups, will be essential to ensure that technology-enhanced adherence solutions are accessible, sustainable, and scalable across diverse healthcare settings.

\subsection*{Limitations}

The current study is based on a small but racially and ethnically diverse sample of female BCS who were, on average, middle-aged and diagnosed with early-stage breast cancer. Approximately half of the patients received patient navigation. Exploring the generalizability of our approach in larger and more diverse patient populations with other types of chronic conditions and medication regimens might be beneficial. Furthermore, our approach relies on the assumption that opening the smart pill bottle estimates actual medication taking. While using smart pill bottle data as a proxy for adherence has its limitations, it offers a more detailed and continuous measure of adherence behavior compared to traditional methods widely used in literature, such as self-reported surveys and pharmacy refill records, which often lack granularity and can be prone to recall bias. Some researchers have explored the use of camera recordings to verify medication ingestion directly; however, this method is highly intrusive and impractical for real-world adherence monitoring. Other potential sources of bias in our findings include the Hawthorne effect (where people change their behavior when they know that they will be observed) and using self-reported survey data. To address the potential for a Hawthorne effect, we examined the MEMS data for unusual patterns at the beginning of data collection. Ultimately, we excluded the first month of MEMS device data as many participants had very high adherence or had data indicating that they were adjusting to using the MEMS device (e.g., missing data for a week beginning at the time of data collection). Another limitation is the reliance on self-reported survey data, which introduces the possibility of recall bias and social desirability bias. For instance, a patient might wrongly remember how much he/she was bothered by a side-effect of the medication when asked at 4-month intervals. Or she may report more positive attitudes towards medicines to appear more socially desirable. Furthermore, we have a limited understanding of context as it is solely based on fine-grained pill bottle opening events data. We are currently deploying a system consisting of sensor-rich smartphones, wireless MEMS, and wearable sensors that collect in-situ objective data and will mitigate the impact of these limitations and shed more light on the specific context of adherence patterns. However, the widespread implementation of these technology-based solutions could face barriers related to cost, technological literacy, and availability of infrastructure. Future research should explore strategies to overcome these barriers, such as offering alternative methods for monitoring adherence or adapting the technology to be more user-friendly and cost-effective to ensure the inclusivity of the intervention across diverse patient populations.

\section*{Conclusion}
In this work, we develop theory-guided, multi-scale machine learning models to demonstrate how dynamic and static data can be leveraged to identify patients who are at high risk for non-adherence. We are currently expanding on this research to better understand medication-taking behaviors among breast cancer survivors through the collection of multiple streams of data (i.e., survey, ecological momentary assessment, MEMS, wearable sensors, phone sensors) over six months' time. The deployment of this system of data collection will allow us to evaluate other predictors of medication adherence and also allow us to determine which factors may not be important in predicting medication adherence over various timeframes (e.g., daily, weekly). Findings from data collected from this study will be used to develop personalized intervention approaches that can be deployed to people based on their particular risk factors for medication non-adherence via a mobile smartphone application (e.g., tailored reminders, provider alerts, and behavioral support strategies). Following the development of the intervention, we will conduct a future study that both collects the different streams of data and deploys the intervention to evaluate its feasibility and acceptability among BCS. We plan to collect data regarding implementation barriers, such as difficulty using the various types of technology, through these two studies and ultimately hope to reduce the burden of data collection by identifying key factors (e.g., key surveys or survey items) associated with medication adherence. By identifying the most critical adherence predictors, healthcare providers could integrate risk stratification models into clinical workflows to proactively identify patients in need of additional support.

\section*{Supporting information}

\paragraph*{S1 Appendix.}
\label{S1_Appendix}
{\bf Details of surveys and their scoring} This file contains information on different surveys used in our analysis with their description and scoring methods.

\section*{Acknowledgments}
We would like to thank all of the BCS who participated in the study.

\nolinenumbers

%
%
%






\end{document}